%% file: main.tex
\documentclass[10pt,twocolumn,letterpaper]{article}

\usepackage{cvpr}              

\usepackage{multirow}

\input{preamble}

\definecolor{cvprblue}{rgb}{0.21,0.49,0.74}
\usepackage[pagebackref,breaklinks,colorlinks,citecolor=cvprblue]{hyperref}

\def\paperID{*****} 
\def\confName{CVPR}
\def\confYear{2024}

\title{Key Patches Are All You Need: A Multiple Instance Learning Framework For Robust Medical Diagnosis}

\author{D. J. Ara\'ujo$^{1}$, M. R. Verdelho$^{1}$, A. Bissoto$^{2}$, J. C. Nascimento$^{1}$, C. Santiago$^{1}$, C. Barata$^{1,3}$\\
$^{1}$Institute for Systems and Robotics, LARSyS, Instituto Superior T\'ecnico, Portugal\\
$^{2}$Institute of Computing, Recod.ai Lab, University of Campinas, Brazil \quad\quad $^{3}$Lisbon ELLIS Unit\\
{\tt\small diogoparaujo@tecnico.ulisboa.pt }}

\begin{document}
\maketitle
\input{sec/0_abstract}    
\input{sec/1_intro}

\input{sec/2_background}

\input{sec/3_approach}
\input{sec/4_exp_setup}

\input{sec/5_exp_results}

\input{sec/6_conclusion}

{\small
\section*{Acknowledgements}

This work was supported by LARSyS funding (DOI: 10.54499/LA/P/0083/2020, 10.54499/UIDP/50009/2020, and 10.54499/UIDB/50009/2020) and projects 2023.02043.BDANA, 10.54499/2022.07849.CEECIND/CP1713/CT0001, MIA-BREAST [10.54499/2022.04485.PTDC], PT SmartRetail [PRR - C645440011-00000062], Center for Responsible AI [PRR - C645008882-00000055].
}

{
    \small
    \bibliographystyle{ieeenat_fullname}
    \bibliography{main}
}

\input{sec/X_suppl}

\end{document}

%% file: preamble.tex
\usepackage[dvipsnames]{xcolor}


%% file: sec/0_abstract.tex
\begin{abstract}
Deep learning models have revolutionized the field of medical image analysis, due to their outstanding performances. However, they are sensitive to spurious correlations, often taking advantage of dataset bias to improve results for in-domain data, but jeopardizing their generalization capabilities. In this paper, we propose to limit the amount of information these models use to reach the final classification, by using a multiple instance learning (MIL) framework. MIL forces the model to use only a (small) subset of patches in the image, identifying discriminative regions. This mimics the clinical procedures, where medical decisions are based on localized findings. We evaluate our framework on two medical applications: skin cancer diagnosis using dermoscopy and breast cancer diagnosis using mammography. Our results show that using only a subset of the patches does not compromise diagnostic performance for in-domain data, compared to the baseline approaches. However, our approach is more robust to shifts in patient demographics, while also providing more detailed explanations about which regions contributed to the decision.
Code is available at: \href{https://github.com/diogojpa99/Medical-Multiple-Instance-Learning}{https://github.com/diogojpa99/Medical-Multiple-Instance-Learning}.
\end{abstract}

%% file: sec/1_intro.tex
\section{Introduction}
\label{sec:intro}
Deep learning (DL) architectures revolutionized the field of medical image analysis, achieving performances that rival even those of more experienced clinicians. It is undeniable that DL models can extract relevant and sometimes new information from medical data. However, there is still a high degree of uncertainty associated with the information that is being used by these models and whether it maps to actual (novel) concepts, or if the models are identifying spurious correlations and taking advantage of dataset bias \cite{geirhos2020,bissoto2024}. Thus, in order to really leverage DL systems in healthcare, it is necessary to ensure that these models are simultaneously explainable and able to achieve good performances outside the datasets they were trained on.

The evolution in the DL field has led to the proposal of different ways for extracting information from images. In this scope, convolutional neural networks (CNNs) are still the most common architectures in medical image analysis. However, in recent years, vision transformers (ViTs) have also gained popularity \cite{chen2022,azad2023}. CNNs and ViTs adopt different feature extraction paradigms: CNNs explore the local neighborhood, while ViTs are able to capture the image context by using self-attention blocks that leverage spatial information and distant relationships. ViTs also adopt an explicit patch-based strategy, as opposed to the traditional full-image analysis performed by CNNs. Nevertheless, both architectures end up learning patch-based representations that are then aggregated into a single representation vector (\textit{e.g.}, through the global average pooling operation in CNNs and the class token in ViTs). 

Patch or region-based analysis resembles clinical practice for medical image inspection, where doctors search for localized findings and criteria to perform a diagnosis. However, contrary to DL models, clinicians do not need to process all regions in a medical image, as they are able to automatically identify the key regions that match a malignant diagnosis. An example is the 7-point checklist method used in skin image analysis \cite{Leo_2004}. This approach focuses solely on the presence or absence of certain dermoscopic features within the lesion, regardless of their spatial arrangement. Another example is breast cancer, where radiologists identify and classify findings in mammography.

The development of DL models capable of identifying regions of interest (ROIs) in medical images and using only those regions to perform a diagnosis is a promising line of research. On one hand, these methods are more aligned with clinical practice. Additionally, showing the ROIs grants some measure of explainability to the model. On the other hand, by forcing the model to use only a part of the image to perform a diagnosis, we can: i) improve its robustness to bias, as the information that the model can use is limited and thus it must select the most discriminative one; and ii) identify spurious correlations learned by the model (\textit{e.g.}, one or more ROIs matching artifacts instead of clinical findings).

The multiple instance learning (MIL) framework, commonly used in weakly-supervised problems, emerges as a natural direction to enforce CNNs and ViTs to look for ROIs. Under the MIL framework, an image is considered a 'bag', and each patch within the image is an 'instance'. The classification of the entire image depends on the presence or absence of 'key instances', where we can limit their number to be small, forcing the model to make a decision with less information. 

In this paper, we explore the benefits of incorporating a MIL framework on top of the feature extraction procedures of both CNNs and ViTs. Using as test beds two medical problems (skin and breast cancer) we show that MIL can be easily integrated into the pipelines of both CNNs and ViTs and that it can be used to select the most relevant patches for both approaches, reducing the amount of information used by the classifier. The models that use MIL achieve competitive performances against the standard CNNs and ViTs, showing that discriminative information is localized in a (small) subset of image regions. Moreover, by identifying these regions, we can provide the user with explanations for the model's decision. Surprisingly, we also observed that, by using MIL, we obtain diagnostic systems that generalize better to new datasets, with different distributions and characteristics than those used for model training.

The rest of our paper is organized as follows. In Section \ref{sec:rel_work}, we discuss the application of patch selection methods in medical image analysis and how MIL-based approaches can be used in this context. Our approach, which focuses on incorporating various MIL frameworks after the feature extraction pipeline of a CNN or ViT, is discussed in Section \ref{sec:architecture}. The experimental setup and the results are described in Sections \ref{sec:expsetup} and \ref{sec:results}. Finally, our conclusions and findings are summarized in Section \ref{sec:conclusion}.

%% file: sec/2_background.tex
\begin{figure*}[t]
    \centering
    \includegraphics[width=.7\linewidth]{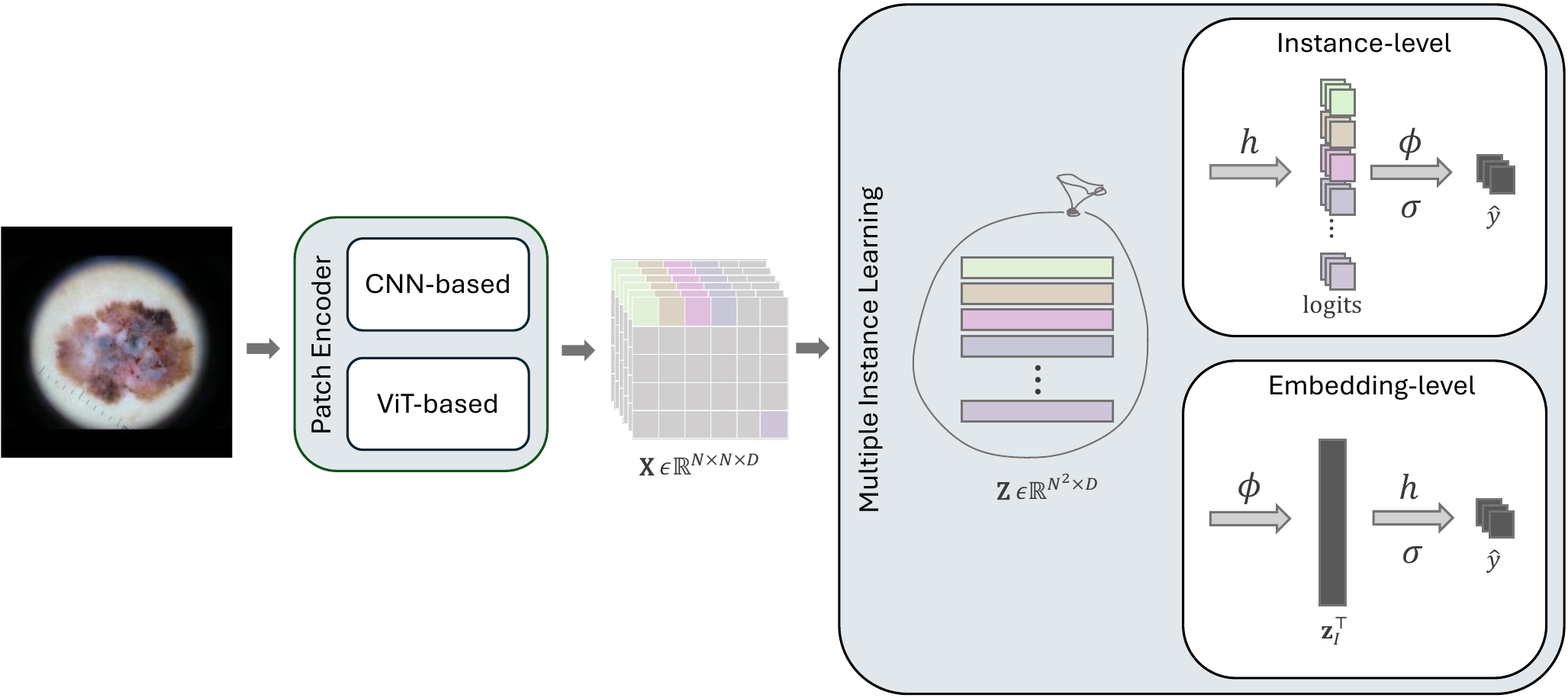}
    \caption{Overview of the proposed approach. An encoder block (CNN or ViT-based) extracts patch representations from the input image. Each patch will be an instance of a bag. Then, a MIL block determines the bag label using an \textit{instance} or \textit{embedding-level} approach.}
    \label{fig:proposal_architecture}
\end{figure*}

\section{Related Work}
\label{sec:rel_work}

Most state-of-the-art classification models for medical image analysis are either based on CNNs or ViTs \cite{chen2022,azad2023}. While these architectures are conceptually different, both can be viewed as extractors of patch-level features. These features are then aggregated into a single vector that represents the entire image. Pooling operators, in particular global average pooling, are usually used for CNNs, while ViTs integrate the information of all patches into the class token. In the end, the representation is fed to an MLP head that performs the binary or multi-class classification.
This means that, when training a model, we are allowing it to explore all the available information to define the decision boundaries. The drawback is that the model can learn spurious correlations and use them to achieve higher performances during training and in-domain validation \cite{geirhos2020,zhang2021}.

Attention blocks, in particular spatial ones, can be seen as a mechanism to reduce the amount of image information used by the models \cite{gonccalves2022}, as they act as patch selectors. Despite their popularity, spatial attention blocks are not sparse (apart from a few exceptions~\cite{martins2021sparse}), which means that all regions in the image end up contributing to the decision.
Moreover, they consist of additional layers of parameters to be learned end-to-end, increasing the model complexity, and their placement in the architecture is not trivial. 

A particular type of attention is self-attention used by ViTs \cite{Dosovitskiy_2020}. Here, multi-head self-attention (MSA) blocks~\cite{Vaswani_2017_v2} are used to extract complex features by leveraging patch correlations. Self-attention also affects the class embedding that will be fed to the classifier. However, while the visualization of the output of the MSA blocks allows the identification of relevant patches, they are all used to build the class embedding. To overcome this issue and reduce the amount of information, Liang \textit{et al.} \cite{Liang_2022} introduced the Expediting ViT (EViT), which progressively discards less relevant patches.
The EViT model uses attention scores to determine the significance of each patch towards the model's output. The $k$ most relevant patches are categorized as ``attentive'', while the remaining ones are deemed ``inattentive'' and are subsequently merged into a single embedding. EViT showcases a promising direction for information selection in ViTs. However, the ratio between attentive and inattentive patches must be empirically defined by the user.

MIL-based frameworks have been explored in the field of medical image analysis to process high-resolution images, such as those of histopathology \cite{Ilse_2018,shao2021,li2021,chen2024}. MIL is particularly well suited in this context, as often we only have access to image level labels (\textit{e.g}, tumor staging), but the relevant information is localized in a small portion of the image that we want to identify. To achieve this goal, the original image is partitioned into big patches that are then independently processed by a CNN and aggregated in the end, using a variety of strategies such as max or attention pooling \cite{Ilse_2018} and transformers \cite{shao2021}. 

It is clear that MIL frameworks are explainable, as they highlight relevant patches in an image. Moreover, when certain operators are used (max or top-k pooling) they can be seen as a proxy to a spatial attention module that does not require the learning of additional model parameters. However, the application of MIL in the medical domain has two limitations: i) it is often applied to binary classification problems, while several medical problems are multi-class (\textit{e.g.}, skin image analysis); and ii) by dividing the image into patches and processing each one independently, we may be losing relevant features.
Regarding the latter issue, we propose to apply MIL only after the feature extraction processes of CNNs and ViTs. This allows us to select relevant patches and reduce the amount of information used by the classifier, while still exploring the capabilities of these two architectures to extract information from images. For the multi-class problem, there have been some attempts to extend MIL to this setting~\cite{Xu_2007, Pathak_2014, Li_2018}, without definitive results. In this work, we propose a generalized MIL formulation for a multi-class problem and show that the binary problem is a particular case of this formulation.

%% file: sec/3_approach.tex
\section{Proposed Approach}\label{sec:architecture}

Our approach, illustrated in Fig.~\ref{fig:proposal_architecture}, relies on MIL strategies to obtain the image classification using only a subset of its patches. The input image is processed with an encoder block that extracts patch features. Then, a MIL block predicts the classification of the image (bag), based on those patch (instance) representations. The following sections describe each block in detail.

\subsection{Patch Encoder Block}

The first component of our approach is a patch encoder block. This block is responsible for generating the representation vector of each of the $N\times N$ patches in the image. Two types of encoders may be adopted: CNNs, where patch representations correspond to each pixel of the output feature map; or ViTs, where patch representations correspond to the final representation of each patch token.

\subsubsection{CNN Encoder}

Most popular CNNs follow the same type of architecture, consisting of a sequence of convolutional blocks (with convolutional, pooling, and normalization layers), followed by a classification head. The convolutional blocks process the image using small kernels that extract low-level to high-level features from each region in the image. Their output is a $N\times N$ feature map, $\mathbf{X} \in \mathbb{R}^{N \times N \times D}$, with $N$ much smaller than the size of the original image, illustrated in Fig. \ref{fig:proposal_architecture}. 

Due to the convolution operations, each pixel in this feature map can be interpreted as the representation of a patch (receptive field) in the image. Typically, the feature map is then transformed using a global average pooling, resulting in a representation vector for the entire image that is the input to the classification head. The underlying premise of this step, however, is that the relevant information for the classification task is spread across the entire image. 

To avoid the above premise, we discard the global average pooling and treat each pixel in the feature map as an instance for the MIL classifier. Concretely, we assume that the feature map $\mathbf{X}$ contains the representation of all $N\times N$ patches in the image.

\subsubsection{ViT Encoder}

ViT-based architectures use the multi-head self-attention (MSA) mechanism~\cite{Vaswani_2017} to extract complex features based on patch correlations in images while taking into account positional information. The input image is first transformed into a sequence of $N^2$ patches. Then, these patches are processed by several linear projections and MSA layers. 

Each MSA consists of running several self-attention mechanisms in parallel on a sequence comprising the patch representations and an additional class token. The resulting attention maps hold information regarding the pairwise similarities between patches. Effectively, each MSA layer modifies the patches and the class token representations by combining the information contained in the entire sequence through weighted averages.

In the standard ViT, the final image classification is obtained by applying an MLP head to the class token, which harnesses information from all patches. To avoid this, we discard the class token and use the final patch representations, denoted as $\mathbf{X} \in \mathbb{R}^{N \times N \times D}$, as input to the MIL block. This means that each patch representation captures the global context of the image, unlike in the CNN encoder, where they only depend on their local neighborhood.

\subsection{MIL Block}

The MIL block aims to apply a MIL classifier to the patch representations obtained by the patch encoder block. The tensor $\mathbf{X}$ is first flattened to a matrix $\mathbf{Z} \in \mathbb{R}^{N^2 \times D}$, which is our collection of instances, represented as a bag in Fig. \ref{fig:proposal_architecture}. The $j$-th line vector in $\mathbf{Z}$, $\mathbf{z}_j \in \mathbb{R}^{1 \times D}$, is the representation of the $j$-th patch in the input image. 
The proposed MIL classifier consists of three key operations:
\begin{itemize}
    \item $h$, a linear projection function that maps the input from an embedding of dimension $D$ to the logits with dimension $C$ (the number of classes), given by $h(\textbf{Z}) = \textbf{W}\textbf{Z}^\top + \textbf{b}$, where $\mathbf{W} \in \mathbb{R}^{C\times D}$ and $\mathbf{b}\in \mathbb{R}^C$ is a bias term;
    
    \item $\phi$, a permutation-invariant pooling that aggregates the patch-level information into image-level by applying a top-$k$ average, where $k=1$ leads to max pooling and $k=N^2$ leads to average pooling; and
    
    \item $\sigma$, a non-linear activation function.
\end{itemize}
The order of these three functions, $\{ h, \phi, \sigma \}$, determines the specific MIL approach used: \textit{instance-level} or \textit{embedding-level}, as detailed in the following sections. 

This formulation is a generalization of the classical MIL approach for binary problems. However, it should be emphasized that the binary case represents a special case of our formulation, where we set $C=1$ and $\sigma$ is the sigmoid function. On the other hand, in a multi-class problem, $C>2$ and $\sigma$ is the softmax function.

\subsubsection{Instance-level Approach}

The \textit{instance-level} approach is characterized by performing class predictions on each patch. It can be implemented in two different modes, although in both cases the first step is to apply $h$ to the patch representations. This projects the patch features into a new embedding space of dimension $C$, corresponding to the patch logits.

For the first \textit{instance-level} mode (I-1), the second step is to apply $\sigma$, which converts the logits obtained in the previous step into class probabilities for each patch. Then, in the final step, the pooling function $\phi$ is applied, resulting in the predicted class probabilities $\hat{y} \in [0,1]^C$.

The second \textit{instance-level} mode (I-2) reverses the order of these two steps. It applies the pooling $\phi$, followed by $\sigma$, to convert the pooled logits into probabilities.

Notice that in the special case of a binary problem, I-1 and I-2 lead to the same classification result, even though they estimate different probabilities for the two classes. Therefore, we only show results with I-1 for this setting.

\subsubsection{Embedding-level Approach}

The \textit{embedding-level} approach starts by aggregating patch representations with the pooling function, $\phi$. This leads to a new vector, $\textbf{z}_I \in \mathbb{R}^{1\times D}$, representing image-level features. Only then are these features transformed to logits with the linear projection $h$. Finally, the $\sigma$ operator converts the logits to the predicted class probabilities, $\hat{y} \in [0,1]^C$.

When $\phi$ is the average pooling, this approach reverts back to the standard CNN strategy of applying a global average pooling before the classification head.

%% file: sec/4_exp_setup.tex
\section{Experimental Setup}\label{sec:expsetup}

We evaluated the performance of the proposed approach in two medical image classification problems: skin cancer diagnosis in dermoscopy and breast cancer diagnosis in mammography.
For each of these settings, we trained a set of baselines (standard CNNs and ViTs models) and our MIL approaches described in Section \ref{sec:architecture}. In order to compare with a recent approach that also performs patch selection, we trained various EViTs \cite{Liang_2022} with different keep rate values for the attentive patches. Our results are all evaluated in terms of class recall (R) and balanced accuracy (BA - the average of the recalls). Below, we describe the adopted datasets, as well as the training specifications. 

\subsection{Datasets}
\textbf{Skin Cancer}.
For dermoscopy image analysis, we address two main challenges: binary and multi-class classification. The ISIC 2019 dataset~\cite{Codella_2016, Tschandl_2018, Combalia_2019} is our primary dataset, which we partition into a training ($80\%$) and validation ($20\%$) sets. For the binary problem, the training phase consisted of using only the melanoma (MEL) and nevi (NV) classes from the ISIC 2019. We also evaluated the generalization capabilities of the proposed approach in several out-of-domain datasets: HIBA~\cite{Lara_2023}, $\mathrm{PH}^2$~\cite{Mendonca_2013}, and  Derm7pt~\cite{Kawahara_2019}.
Each of the previous datasets contains images collected from patients of different demographic groups, allowing us to do a preliminary assessment of the fairness of the different models. 

For the multi-class classification task, we employed the ISIC 2019 dataset~\cite{Codella_2016, Tschandl_2018, Combalia_2019} for training and validation purposes, while the HIBA dataset~\cite{Lara_2023} served as our testing ground. These datasets encompass eight diagnostic categories: Actinic keratosis (AK), Basal cell carcinoma (BCC), Benign keratosis (BKL), Dermatofibroma (DF), Melanoma (MEL), Melanocytic nevus (NV), Squamous cell carcinoma (SCC), and Vascular lesion (VASC). Table~\ref{tab:skin_datasets_class_distribution} provides a detailed overview of the class distributions for the training, validation, and test datasets for both binary and multi-class scenarios.

\begin{table}[t]
    \centering
    \caption[Summary of the overall distribution of the training, validation, and testing datasets.]{Summary of the overall distribution of the training, validation, and testing dermoscopic image datasets.}
    \small
    \label{tab:skin_datasets_class_distribution}
    \begin{tabular}{l cc c c c}
        \toprule
        \multicolumn{1}{c}{\multirow{2}{*}{Classes}} & \multicolumn{2}{c}{\textbf{ISIC 2019}} & \multicolumn{1}{c}{\textbf{HIBA}} & \multicolumn{1}{c}{\textbf{$\mathbf{PH^2}$}} & \multicolumn{1}{c}{\textbf{Derm7pt}} \\
        \cmidrule(lr){2-3} \cmidrule(lr){4-4} \cmidrule(lr){5-5} \cmidrule(lr){6-6}
         & {Train} & {Val.} & {Test} & {Test} & {Test} \\ 
        \midrule
        AK & 687 & 173 & 46 & --- & --- \\
        BCC & 2,653 & 664 & 228 & --- & --- \\
        BKL & 2,089 & 525 & 62 & --- & --- \\
        DF & 191 & 48 & 39 & --- & --- \\
        MEL & 3,611 & 904 & 194 & 40  & 252 \\
        NV & 10,293 & 2,575 & 549 & 160 & 575 \\
        SCC & 502 & 126 & 111 & --- & --- \\
        VASC & 202 & 51 & 41 & --- & --- \\
        \midrule
        \textbf{Total} & 20,228 & 5,066 & 1,270 & 200 & 827\\
        \bottomrule
    \end{tabular}
\end{table}


\textbf{Breast Cancer}.
For mammography image analysis, we evaluated our proposal on the binary task of distinguishing breasts with findings from those with no findings.
We employed the DDSM dataset~\cite{Heath_1998, Heath_2007} for both training (90\%) and validation (10\%). Specifically, the training dataset comprised $2,428$ cases identified with findings and $1,342$ cases without findings. For validation, we evaluated $260$ cases with findings against $137$ cases without findings. 

\textbf{Preprocessing}.
All input images were resized to a uniform size of $224 \times 224 \times 3$. Mammography images were converted from grayscale to RGB by replicating the color channel. To preserve the original aspect ratio of both the dermoscopy and mammography images, we applied padding to ensure that all images had a square format.

\subsection{Training Setup}
\textbf{Encoder Block}. We explored a variety of CNN-based pre-trained backbones for the patch encoder block, encompassing ResNet-18 (RN-18), ResNet-50 (RN-50), VGG-16, DenseNet-169 (DN-169), and EfficientNetB3 (EN-B3). Additionally, we explored ViT models -- DEiT-S and EViT-S with a keep rate ($K_r$) of $0.7$. Every model was pre-trained on the ImageNet1k dataset~\cite{deng2009imagenet}.

For ViT-based encoders, images were partitioned into $14\times 14$ patches. To match these resolutions in the CNN experiments, we collected the CNN feature maps with a spatial dimension of $14\times14$. Among the evaluated CNN backbones, EN-B3 emerged as the best model, leading to the creation of the MIL-EN-B3 model with $2.2\mathrm{M}$ parameters. From the transformer variants, DEiT-S was chosen, forming the MIL-DEiT-S model with $22\mathrm{M}$ parameters. Comparison between the various backbones can be seen in Supplementary Material. 

For the \textit{instance-level} approach, we conducted experiments with three MIL pooling operators: the \textit{max} operator, the \textit{average} operator, and the \textit{top-$k$ average} operator, with three different configurations: $k\approx12.5\%$, $k=25\%$, and $k=50\%$. Our experiments showed that $k=25\%$ was generally the best representation of the \textit{top-$k$ average} pooling operator. In the case of the \textit{embedding-level} approach, we used the following MIL pooling operators: the \textit{column-wise global max} pooling operator, the \textit{column-wise global average} pooling operator, and the \textit{column-wise global top-$k$ average} pooling operator, with $k=25\%$ (different $k$ values were tested and can be seen in Supplementary Material). 

\textbf{Baseline Models}. The baseline models for our experiments used the same backbones as the MIL models described above. Here, however, we use the full architecture, replacing only the classification layer with one specific to our medical problems. 

\textbf{EViT Baselines}. We adopted the EViT-S configuration with $22.1\mathrm{M}$ parameters as the standard for comparing our method. In all configurations, we placed the token reorganization block in three different layers: the $3^{rd}$, the $6^{th}$, and the $9^{th}$ layers. The keep rate ($K_r$) determines the number of attentive tokens retained by the token reorganization block. We explored different settings and settled on $K_r=0.6$ and $K_r=0.7$. With these choices, the EViT model with $K_r=0.6$ preserves $43$ patches, while the model with $K_r=0.7$ retains $68$ patches out of $196$ patches. The assessment of additional $K_r$ values can be seen in the Supplementary Material.

\textbf{Training Configurations}. All models were trained using a class-weighted categorical Cross Entropy (CE) loss function since all datasets are highly unbalanced. Online augmentation strategies tailored to each task were used in order to enhance model robustness. Specifically, for dermoscopy image classification, we adopted the augmentation configuration outlined by Touvron \textit{et al.}~\cite{touvron2021training}. In contrast, the mammography image classification task incorporated random horizontal and vertical flips, along with random rotations. 
All tested models were implemented and trained using PyTorch on NVIDIA GeForce RTX 3090 and 4090.

%% file: sec/5_exp_results.tex
\section{Experimental Results} \label{sec:results}
The experimental results for the binary problems can be seen in Tables \ref{tab:binary_skin_and_breast_val_results} and \ref{tab:skin_generalization_results}, where the latter table corresponds to the generalization experiments. The multi-class results are shown in Table \ref{tab:mutliclass_skin_results}. In the following subsections, we discuss the experimental results as well as the visualizations obtained with our MIL models.

\subsection{Binary MIL}
Our experimental results for the binary classification of dermoscopy and mammography images are summarized in Table~\ref{tab:binary_skin_and_breast_val_results}. These results show that there is a marginal difference between the MIL models and their baseline counterparts. For the ISIC 2019 set, the standard deviation for BA stands at a modest $1.60\%$, and for the DDSM dataset, an even smaller standard deviation of $0.49\%$ is observed. In terms of backbones, the one based on DEiT-S achieves better performances in the case of skin cancer, suggesting that in this context the patch correlation may contain discriminative information. In the case of breast cancer, it seems that the performances are fairly similar. When comparing the DEiT-S and MIL-DEiT-S results with those of EViT, we conclude that: i) discarding several patches does not significantly affect the performance of the models; and ii) our MIL framework is very competitive against more complex models for information selection. Finally, regarding MIL with instances against the embedding versions, we conclude that performing an analysis at the patch level seems to be better in most settings.


\begin{table}[t]
\centering
\caption{Results for the binary problem in dermoscopy and mammography.}
\label{tab:binary_skin_and_breast_val_results}
\addtolength{\tabcolsep}{-1.5pt}
\small
\begin{tabular}{ccc ccc ccc}
\toprule
\multicolumn{3}{c}{\multirow{2}{*}{\textbf{Models}}} & \multicolumn{3}{c}{\textbf{ISIC 2019}} & \multicolumn{3}{c}{\textbf{DDSM}} \\
\cmidrule(rl){4-6} \cmidrule(rl){7-9}
 & & & {BA} & {R-MEL} & {R-NV} & {BA} & {R-F} & {R-NoF} \\
\midrule
\multicolumn{3}{c}{EN-B3} & 90.7 & 85.5 & 95.8 & 96.1 & 94.6 & 97.6 \\
\multicolumn{3}{c}{DEiT-S} & 91.7 & 86.7 & \textbf{96.7} & 95.3 & \textbf{95.0} & 95.6 \\
\midrule
\multirow{2}{*}{\rotatebox[origin=c]{90}{EViT}} 
& \multicolumn{2}{c}{Kr = 0.6} & 91.4 & 86.6 & 96.3 & \textbf{96.2} & 94.6 & 97.8 \\ 
& \multicolumn{2}{c}{Kr = 0.7} & 90.7 & 85.4 & 95.7 & 95.8 & 93.1 & 98.5 \\
\midrule
\multirow{6}{*}{\rotatebox[origin=c]{90}{MIL-EN-B3}} 
& \multirow{3}{*}{\textbf{I}} 
& Max & 88.5 & 86.7 & 90.2 & 95.4 & 91.5 & 99.3 \\
&  & Topk & 89.5 & 85.6 & 93.3 & 95.2 & 91.9 & 98.5 \\
&  & Avg & 89.1 & 86.7 & 91.6 & 94.9 & 92.7 & 97.1 \\
\cmidrule{2-9}
&  \multirow{3}{*}{\textbf{E}} 
& Max & 86.0 & 85.7 & 86.4 & 95.8 & 91.5 & \textbf{100.0} \\
& & Topk & 89.2 & 85.7 & 92.7 & 95.8 & 94.6 & 97.1 \\
& & Avg  & 89.1 & 84.4 & 93.8 & 95.8 & 93.1 & 98.5 \\
\midrule
\multirow{6}{*}{\rotatebox[origin=c]{90}{MIL-DEiT-S}} 
& \multirow{3}{*}{\textbf{I}} 
& Max & 91.7 & 87.1 & 96.3 & 94.7 & 91.5 & 97.8 \\
&  & Topk & 91.4 & 86.6 & 96.2 & 94.9 & 92.7 & 97.1 \\
&  & Avg & \textbf{91.8} & \textbf{87.5} & 96.1 & 95.1 & 93.1 & 97.1 \\
\cmidrule{2-9} 
& \multirow{3}{*}{\textbf{E}} 
& Max & 91.0 & 87.4 & 94.5 & 95.8 & 92.3 & 99.3 \\
& & Topk & 91.5 & 86.9 & 96.1 & 94.7 & 92.3 & 97.1 \\
& & Avg & 91.4 & 87.4 & 95.4 & 95.6 & 91.9 & 99.3 \\
\bottomrule
\end{tabular}
\end{table}

In summary, the binary results underscore the potential of integrating a MIL into CNN and ViT pipelines to select key patches for diagnosis. This process effectively reduces the information used by the classifiers without significant performance loss, suggesting that the most discriminative information is concentrated in a few regions of the images.

\subsection{Multi-class MIL}

In this section, we discuss the results of our proposed multi-class MIL framework, as detailed in Table \ref{tab:mutliclass_skin_results}. The table compares the performance of our MIL methods with that of baseline models and EViT on the challenging task of multi-class classification of dermoscopy images. Here we show the results for ISIC 2019 and HIBA~\cite{Lara_2023}, which was the only test set where all classes matched the ones used for training. In this section, we will only discuss the results for ISIC 2019, while the HIBA results will be discussed in the next section.

\begin{table}[t]
\centering
\caption{Results for the multi-class problem in dermoscopy.}
\label{tab:mutliclass_skin_results}
\small
\begin{tabular}{ccc c c}
\toprule
\multicolumn{3}{c}{\multirow{2}{*}{\textbf{Models}}} & \textbf{ISIC 2019} & \textbf{HIBA} \\
\cmidrule(rl){4-4} \cmidrule(rl){5-5}
 & & & {BA} & {BA} \\
\midrule
\multicolumn{3}{c}{EN-B3} & 82.2 & 32.6 \\
\multicolumn{3}{c}{DEiT-S} & 83.6 & 37.6 \\
\midrule
\multirow{2}{*}{\rotatebox[origin=c]{90}{EViT}} 
& \multicolumn{2}{c}{Kr = 0.6} & 83.6 & 36.2 \\ 
& \multicolumn{2}{c}{Kr = 0.7} & \textbf{84.3} & 36.1 \\
\midrule
\multirow{10}{*}{\rotatebox[origin=c]{90}{MIL-EN-B3}} 
& \multirow{3}{*}{\textbf{I-1}}
& Max & 74.1 & 36.5 \\
& & Topk & 78.4 & 33.3 \\ 
& & Avg & 79.9 & 34.4 \\
\cmidrule{2-5}
& \multirow{3}{*}{\textbf{I-2}}
& Max & 76.4 & 36.3 \\
& & Topk & 76.2 & 33.9 \\ 
& & Avg & 77.5 & 34.9 \\
\cmidrule{2-5}
& \multirow{3}{*}{\textbf{E}}
& Max & 72.3 & 34.9 \\
& & Topk & 78.9 & 35.5 \\ 
& & Avg & 77.6 & 37.1 \\
\midrule
\multirow{10}{*}{\rotatebox[origin=c]{90}{MIL-DEiT-S}} 
& \multirow{3}{*}{\textbf{I-1}}
& Max & 82.2 & \textbf{39.4} \\
& & Topk & 81.7 & 34.5 \\ 
& & Avg & 81.6 & 35.1 \\
\cmidrule{2-5}
& \multirow{3}{*}{\textbf{I-2}}
& Max & 75.4 & 35.4 \\
& & Topk & 79.0 & 33.1 \\ 
& & Avg & 82.6 & 32.9 \\
\cmidrule{2-5}
& \multirow{3}{*}{\textbf{E}}
& Max & 82.4 & 33.8 \\
& & Topk & 82.2 & 35.6 \\ 
& & Avg & 82.6 & 36.2 \\
\bottomrule
\end{tabular}
\end{table}

\begin{table*}[t]
\centering
\caption{Generalization results for binary classification of dermoscopy images.}
\label{tab:skin_generalization_results}
\small
\begin{tabular}{ccc ccc ccc ccc}
\toprule
\multicolumn{3}{c}{\multirow{2}{*}{\textbf{Models}}} & \multicolumn{3}{c}{\textbf{HIBA}} & \multicolumn{3}{c}{$\mathbf{PH^2}$} & \multicolumn{3}{c}{\textbf{Derm7pt}} \\
\cmidrule(rl){4-6} \cmidrule(rl){7-9} \cmidrule(rl){10-12}
 & & & {BA} & {R-MEL} & {R-NV} & {BA} & {R-MEL} & {R-NV} & {BA} & {R-MEL} & {R-NV}  \\
\midrule
\multicolumn{3}{c}{EN-B3} & 81.5 & 68.0 & 94.9 & 88.8 & 82.5 & 95.0 & 76.2 & 57.9 & 94.4 \\
\multicolumn{3}{c}{DEiT-S} & 82.0 & 67.0 & 96.9 & 86.6 & 75.0 & 98.1 & 74.0 & 53.2 & 94.8 \\
\midrule
\multirow{2}{*}{\rotatebox[origin=c]{90}{EViT}} 
& \multicolumn{2}{c}{Kr = 0.6} & 81.5 & 68.0 & 94.9 & 88.8 & 80.0 & 97.5 & 73.4 & 52.8 & 94.1 \\
& \multicolumn{2}{c}{Kr = 0.7} & 82.7 & 71.1 & 94.4 & 86.6 & 75.0 & 98.1 & 74.9 & 56.7 & 93.0 \\
\midrule
\multirow{6}{*}{\rotatebox[origin=c]{90}{MIL-EN-B3}} 
& \multirow{3}{*}{\textbf{I}} 
& Max & \textbf{85.1} & \textbf{76.3} & 93.8 & 84.4 & 75.0 & 93.8 & \textbf{78.7} & \textbf{67.1} & 90.4 \\
&  & Topk & 80.4 & 63.9 & 96.9 & \textbf{89.7} & \textbf{82.5} & 96.9 & 77.0 & 60.7 & 93.2 \\
&  & Avg & 82.7 & 72.7 & 92.7 & 85.0 & 77.5 & 92.5 & 76.5 & 64.3 & 88.7 \\
\cmidrule{2-12}
&  \multirow{3}{*}{\textbf{E}} 
& Max & 85.0 & 74.7 & 95.0 & 85.3 & 77.5 & 93.1 & 76.1 & 66.3 & 85.9 \\
& & Topk & 83.9 & 75.3 & 92.5 & 88.1 & 82.5 & 93.8 & 76.7 & 59.5 & 93.9 \\
& & Avg  & 82.4 & 68.0 & 96.7 & 83.4 & 70.0 & 96.9 & 75.3 & 56.3 & 94.3 \\
\midrule
\multirow{6}{*}{\rotatebox[origin=c]{90}{MIL-DEiT-S}} 
& \multirow{3}{*}{\textbf{I}} 
& Max & 82.1 & 67.5 & 96.7 & 87.2 & 77.5 & 96.9 & 74.4 & 54.8 & 94.1 \\
&  & Topk & 81.0 & 64.9 & \textbf{97.1} & 84.1 & 72.5 & 95.6 & 71.8 & 49.6 & 94.1 \\
&  & Avg & 83.2 & 69.6 & 96.7 & 87.8 & 80.0 & 95.6 & 74.8 & 56.0 & 93.7 \\
\cmidrule{2-12} 
& \multirow{3}{*}{\textbf{E}} 
& Max & 81.3 & 68.6 & 94.2 & 85.3 & 75.0 & 95.6 & 74.7 & 58.3 & 91.1 \\
& & Topk & 83.1 & 71.1 & 95.1 & 89.7 & 80.0 & \textbf{99.4} & 75.1 & 55.2 & \textbf{95.1} \\
& & Avg & 82.2 & 69.6 & 94.7 & 85.0 & 75.0 & 95.0 & 76.0 & 61.1 & 91.0 \\
\bottomrule
\end{tabular}
\end{table*}

When comparing our multi-class MIL framework with the baseline models, we find that the latter performs better in this task. There is a more noticeable difference in performance when using a CNN as the backbone of our model. This disparity might stem from how the EN-B3 model and the MIL model handle feature extraction. Specifically, the EN-B3 model may use different types of features than the MIL model, which extracts feature maps from a previous layer in the network. This leads to models that have a significantly different number of parameters, which may also impact their ability to memorize in-domain features. Specifically, our MIL-EN-B3 model operates with only $2.2\mathrm{M}$ parameters, as opposed to the more substantial $11\mathrm{M}$ parameters of the EN-B3 model.  This hypothesis is further supported by the similar performance between the \textit{instance-level} and \textit{embedding-level} MIL approaches. The \textit{embedding-level} approach mainly serves as a bridge between the MIL framework and traditional CNN-based architectures. The performance comparison between the \textit{embedding-level} MIL-EN-B3 and the EN-B3 baseline mirrors the gap observed with the \textit{instance-level} MIL, suggesting that the disparities in the results could in fact be attributed to the different sizes of the model architectures.

Regarding the two \textit{instance-level} approaches for multi-class classification (I-1 and I-2), it seems that I-1 performs better across the two backbones. This leads us to recommend the I-1 formulation in future multi-class MIL applications. Once more, the \textit{instance-level} MIL seems to consistently outperform the embedding approach, reinforcing the importance of performing a patch-based analysis rather than aggregating all or a subset of the image information.

The multi-class classification task is inherently more challenging than its binary counterpart. Nevertheless, the results from the EViT model still prove that information selection is desirable and leads to improved performance compared to traditional models that classify over the entire image. Moreover, our MIL models with DEiT-S backbone still hold their own against both the baselines and EViT. These findings challenge the assumption that larger, more complex models are always better. In essence, our results argue for a more targeted, efficient approach to medical image analysis.

\subsection{Generalization Across Diverse Demographies}

\begin{figure*}[t]
    \centering
    \includegraphics[width=.8\linewidth]{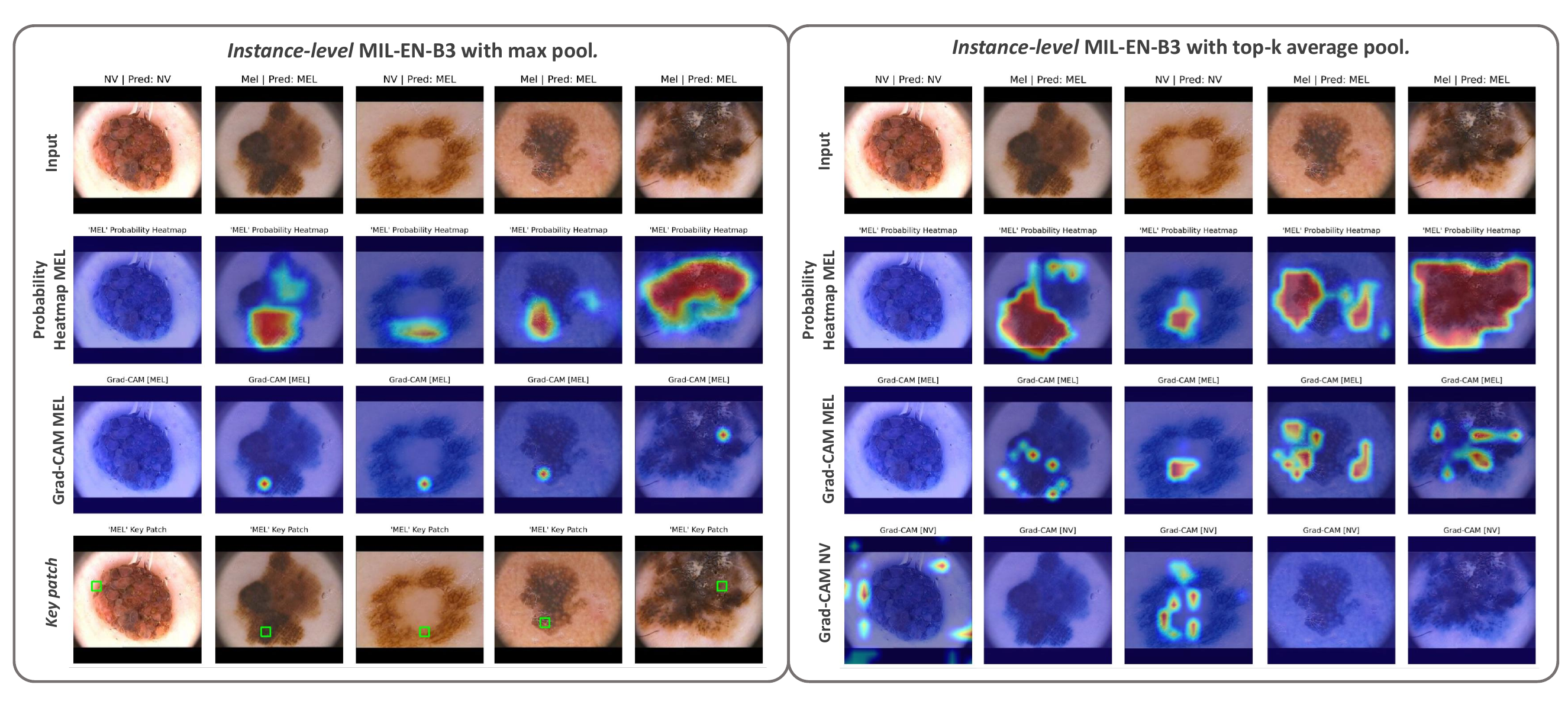}
    \caption{Visualization of the key patches identified by two different MIL approaches. On the left, we have the \textit{instance-level} MIL-EN-B3 using \textit{max} polling, and on the right, we have the \textit{instance-level} MIL-EN-B3 using the \textit{top-$k$ average} pooling operator, with $k\approx12.5\%$, The images used for visualization are taken from the $\mathrm{PH}^2$ test set and refer to the binary classification task of melanoma vs. nevus.}
    \label{fig:results_mil_inst_max_vs_topk}
\end{figure*}
\begin{figure}[t]
    \centering
    \includegraphics[width=.8\linewidth]{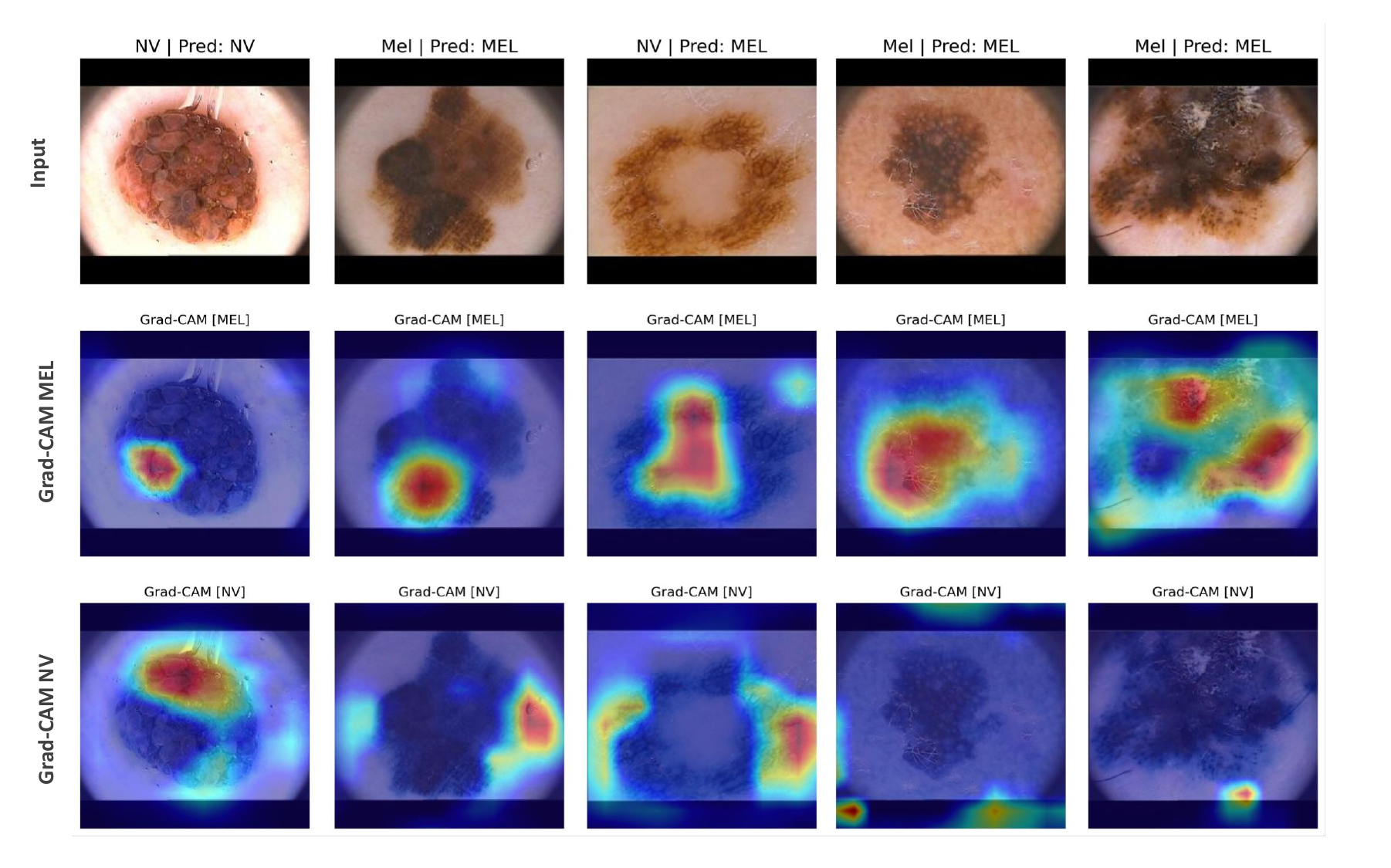}
    \caption{Grad-Cam \cite{Shah_null} heatmap visualizations generated by the EN-B3 baseline model for images from the $\mathrm{PH}^2$ test set.}
    \label{fig:cnn_gradCam}
\end{figure}

We evaluated the robustness of our MIL-based models across varied dermoscopy image datasets, each representing distinct patient demographics not seen in our training or validation sets. Table \ref{tab:skin_generalization_results} displays the binary results for skin tests from the HIBA, $\mathrm{PH}^2$, and Derm7pt datasets, while Table~\ref{tab:mutliclass_skin_results} shows the results for HIBA.

Our MIL models consistently outperformed their baseline counterparts on unseen data. For example, the \textit{instance-level} MIL-EN-B3 model using max pooling outperformed the EN-B3 baseline by a BA of $3.6\%$ on the HIBA dataset. In addition, the multi-class results on the HIBA test set (see Table \ref{tab:mutliclass_skin_results}) show that even if there is a performance drop when the MIL models are compared with the baselines, they still generalize better to unseen data. Finally, contrary to what is stated in the literature, the \textit{embedding-level} approach did not consistently outperform the \textit{instance-level} models. In fact, the \textit{instance-level} models often outperformed their \textit{embedding-level} counterparts. These results suggest that the key patches identified by the \textit{instance-level} MIL models may have significant medical relevance, contributing to improved generalization to unseen data.

In summary, the generalization results show that our MIL models deal better with unseen data, being potentially more fair across different demographics, despite using less information. This establishes MIL as a promising method to improve fairness in medical image analysis. 


\subsection{Explainability of MIL Models}
Our results suggest that the key regions identified by the \textit{instance-level} MIL models are correlated with meaningful information within the image, thereby increasing the model's robustness to dataset bias compared to baseline counterparts. Nevertheless, it is critical to assess whether these identified regions truly capture clinical findings or are simply artifacts. To clarify this, we compared heatmaps generated by the EN-B3 baseline to those generated by our MIL models, focusing on the binary classification of melanoma versus nevus in the $\mathrm{PH}^2$ dataset.

Figure \ref{fig:cnn_gradCam} shows the Grad-Cam visualizations for the baseline EN-B3 model, highlighting the areas that influence its predictions for melanoma (MEL) and nevus (NV) classifications, while Figure \ref{fig:results_mil_inst_max_vs_topk} illustrates the key regions identified by the \textit{instance-level} MIL-EN-B3 model for the same lesions. In the MIL setting, we compare two pooling strategies: \textit{max} pooling and \textit{top-$k$ average} pooling, which averages the values of the 25 most relevant patches. 

Notably, our \textit{instance-level} MIL approaches consistently highlight \textit{key patches} that, similar to ROIs in clinical diagnosis, lie within or at the edges of lesions for melanoma cases, or bordering healthy tissue for nevus cases. This preference for clinically relevant areas confirms the ability of our models to extract medically relevant features, validating their performance on the validation set and their ability to generalize to unseen data.

When comparing the heatmaps, it is clear that the EN-B3 model produces coarse heatmaps, whereas the MIL's \textit{instance-level} approaches produce a more detailed delineation of relevant regions, providing finer explanations. This clarity and specificity reinforces MIL's position as a more clinically translatable tool that can potentially provide clearer explanations of the decision-making process.



%% file: sec/6_conclusion.tex
\section{Conclusions} \label{sec:conclusion}

This work demonstrates the potential of integrating MIL into the pipeline of CNNs and ViTs to select relevant patches and use less information in the classification stage. Our findings reveal that despite a significant reduction in the amount of information, MIL models can achieve results comparable to more complex networks. By focusing on the most discriminative image patches, similar to clinical practice, MIL models show a strong ability to generalize across different datasets and demographic groups. This suggests a promising direction towards creating more explainable, efficient, and fair medical image analysis systems. Moreover, the assessment of selected regions underscores MIL's alignment with clinical relevance, providing a more interpretable decision-making process that mirrors the diagnostic approach of medical experts. Future work will focus on validating the identified key regions against specific medical concepts, as well as exploring these regions to improve model performance and further enhance clinical applicability and fairness.

%% file: sec/X_suppl.tex
\clearpage
\setcounter{page}{1}
\maketitlesupplementary


\section{Additional Results}

\subsection{Top-$k$ Search In The MIL Framework}

\begin{figure}[h]
    \centering
    \includegraphics[width=3.25in]{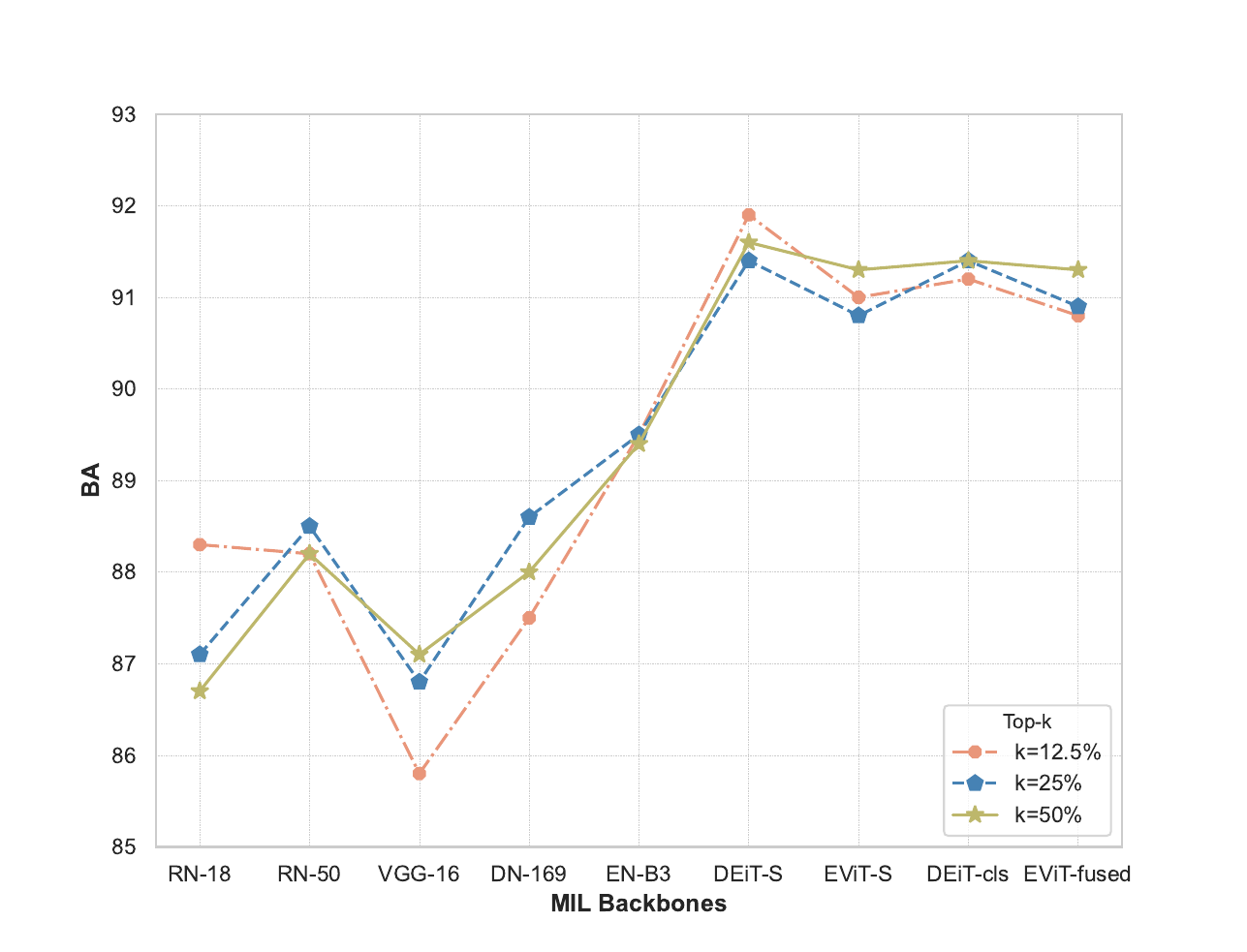}
    \small
    \caption{Search for the optimal $k$ hyperparameter in the \textit{instance-level}\textit{ top-$k$ average} MIL pooling operator. We explored three values for the hyperparameter: $k\approx12.5\%$, $k=25\%$, and $k=50\%$. Our experiments were conducted and evaluated on the validation set of the ISIC 2019 dataset, employing different MIL backbones. The backbones included RN-18, RN-50, VGG16, DN-169, EN-B3, DEiT-S, DEiT-cls (DEiT with the CLS token), EViT-S, and EViT-fused (EViT with the fused embedding). Notably, with EViT backbones, $k\approx12.5\%$ resulted in only $9$ patches, $k=25\%$ retained $17$ patches, and $k=50\%$ maintained $34$ patches. These results indicate that using more patches in the bag evaluation does not necessarily lead to better performance.} 
    \label{fig:mil_top-$k$_search}
\end{figure}

We conducted experiments with three different $k$ values in the \textit{instance-level} approach: approximately $12.5\%$, $25\%$, and $50\%$. In the context of our experiments, which involved input images of $224 \times 224$ resolution and patches of $ 16 \times 16$ resolution, we dealt with a total of $N=196$ patches. This implies that for the $k\approx12.5\%$ configuration, we considered $25$ patches; for $k=25\%$, we worked with $49$ patches, and for $k=50\%$, we used $98$ patches. To determine the optimal $k$ value for the \textit{top-$k$ average} operator in the \textit{instance-level} approach, we conducted experiments using various backbones on the validation set of the ISIC 2019 dataset. A summary of these experiments is shown in Figure \ref{fig:mil_top-$k$_search}.

The plot in Figure \ref{fig:mil_top-$k$_search} shows that the choice of $k$ for the \textit{top-$k$ average} operator in the \textit{instance-level} approach does not significantly impact the performance of the different MIL models. This observation suggests that not all patches within a dermoscopy image contribute equally to the classification task, indicating that the discriminative information lies within a (small) subset of image patches. Interestingly, the $k\approx12.5\%$ and $k=25\%$ scenarios consistently yield the highest BA results across different MIL backbones. Based on these results, we selected $k=25\%$ as the default configuration for the \textit{top-$k$ average} pooling operator. To ensure a fair comparison between the \textit{embedding-level} and \textit{instance-level} approaches, we have also adopted $k=25\%$ as the preferred setting for the \textit{column-wise global top-$k$ average} operator in the \textit{embedding-level} approach.

\subsection{EViT Experimental Setup Complementary Material}

\begin{figure}[h]
    \centering
    \includegraphics[width=3.25in]{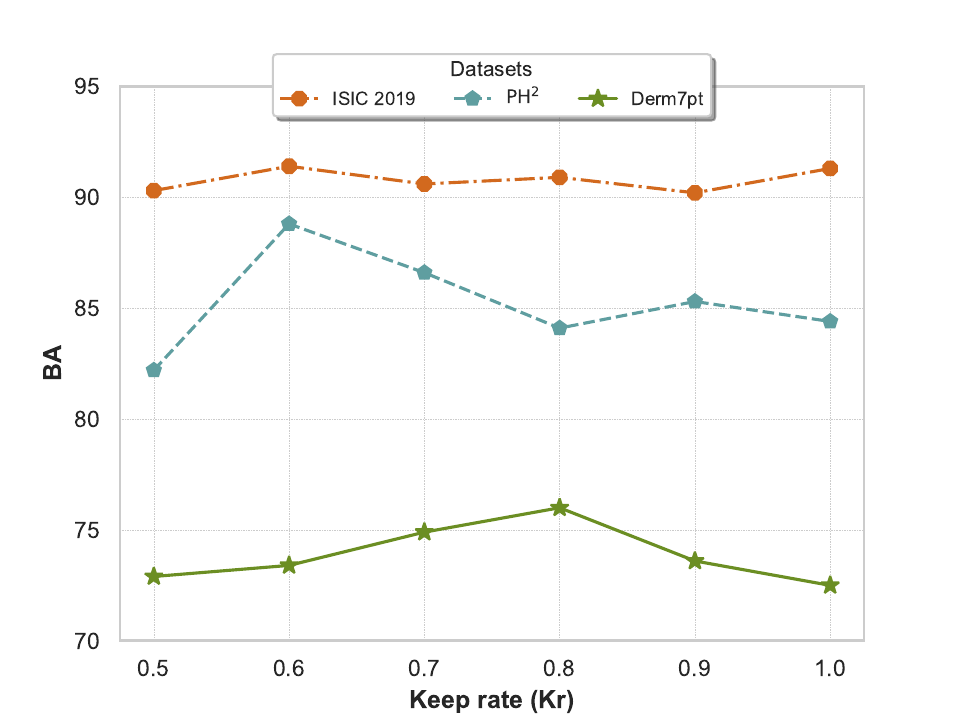}
    \caption{Performance comparison between EViT-S models with different keep rates. The experiments were conducted on the ISIC 2019 validation dataset, as well as on the two test datasets: $\mathrm{PH}^2$ and Derm7pt, for the binary classification task of melanoma versus nevus. The $x$-axis represents the different $K_r$ values, while the $y$-axis represents the corresponding BA results.} 
    \label{fig:EViT_Kr_search}
\end{figure}

The combination of layers in which token reorganization takes place and the choice of $K_r$ (keep rate) are the most critical hyperparameters in the EViT architecture. Since we decided to fix the token reorganization block in the $3^\text{rd}$, $6^\text{th}$, and $9^\text{th}$ layers, an extensive search for the best $K_r$ configurations in the EViT-S model was required. Since the $K_r$ hyperparameter plays a crucial role within the EViT architecture, we conducted a series of experiments to examine its impact on the EViT-S configuration. Figure \ref{fig:EViT_Kr_search} shows the BA results across different $K_r$ values for the ISIC 2019 validation set and both test datasets: $\mathrm{PH}^2$ and Derm7pt. These results indicate that a higher $K_r$ does not necessarily lead to better performance, especially on the test datasets. It is clear that $K_r$ values of $0.6$, $0.7$, and $0.8$ outperform the rest. Therefore, in this work we used the $K_r=0.6$ and $K_r=0.7$ EViT-S configurations.

\subsection{Selection Of MIL Backbones And Baselines}

To facilitate the experiments conducted in this paper, we carefully selected a representative model from each of the CNN and Transformer baselines. This selection process involved an extensive evaluation of each model on the validation set of the ISIC 2019 dataset, focusing on the binary classification problem of melanoma (MEL) versus nevus (NV). The results of this evaluation are summarized in table \ref{tab:baselines_expSetup}. 

\begin{table}[h]
    \centering
    \begin{tabular}{ll ccc}
    \toprule
    \multicolumn{2}{c}{\multirow{2}{*}{Baseline models}} & \multicolumn{3}{c}{\textbf{ISIC 2019}} \\
    \cmidrule(lr){3-5}
     & & {BA} & {R-MEL} & {R-NV} \\ 
    \midrule
    \multirow{5}{*}{\rotatebox[origin=c]{90}{CNN}} 
    & RN-18 & 88.6 & 83.8 & 93.4 \\
    & RN-50 & 88.9 & 82.6 & 95.1 \\
    & VGG-16 & 87.7 & 83.6 & 91.8 \\ 
    & DN-169 & 89.1 & 83.2 & 95.0 \\ 
    & EN-B3 &  \textbf{90.7} & \textbf{85.5} & \textbf{95.8} \\ 
    \midrule
    \multirow{4}{*}{\rotatebox[origin=c]{90}{ViT}} 
    & ViT-S & 91.3 & 86.8 & 95.8 \\
    & ViT-B & 90.6 & 85.3 & 95.8 \\ 
    & DEiT-S & \textbf{91.7} & 86.7 & \textbf{96.7} \\ 
    & DEiT-B & 91.7 & \textbf{87.2} & 96.2 \\ 
    \bottomrule
    \end{tabular}
    \caption{Evaluation results of a set of baseline models on the validation set of the ISIC 2019 dataset. The baseline models include different architectures, including RN-18, RN-50, VGG-16, DN-169, EN-B3 from the CNN-based category, and ViT-S, ViT-B, DEiT-S, DEiT-B from the Transformer-based category. The evaluation is performed for the binary classification problem of melanoma versus nevus.}
    \label{tab:baselines_expSetup}
\end{table}

\subsection{Multi-class MIL Complementary Material}

A detailed summary of the results for the multi-class MIL is shown in Table \ref{tab:mutliclass_skin_results_supp}.

\begin{table*}
    \centering
    \caption{Results of multi-class image classification on the ISIC 2019 validation set. '\textbf{I-1}' and '\textbf{I-2}' denote the first and second \textit{instance-level} approaches, respectively, and '\textbf{E}' denotes the \textit{embedding-level} approach.}
    \label{tab:mutliclass_skin_results_supp}
    \small
    \begin{tabular}{ccc ccc ccc ccc}
        \toprule
        \multicolumn{3}{c}{\multirow{2}{*}{\textbf{Models}}} & \multicolumn{9}{c}{\textbf{ISIC 2019}} \\
         \cmidrule(lr){4-12}
         & & &  {BA} & {R-AK} & {R-BCC} & {R-BKL} & {R-DF} & {R-MEL} & {R-NV} & {R-SCC} & {R-VASC} \\ 
        \midrule
        \multicolumn{3}{c}{EN-B3} & 82.2 & 71.1 & 87.8 & 79.4 & 81.3 & 76.5 & 91.6 & 72.2 & 98.0 \\
        \multicolumn{3}{c}{DEiT-S} & 83.6 & 72.3 & 90.5 & \textbf{82.7} & 87.5 & \textbf{80.3} & 92.1 & 65.1 & 98.0 \\
        \midrule
        \multirow{2}{*}{\rotatebox[origin=c]{90}{EViT}}
        & \multicolumn{2}{c}{Kr = 0.6} & 83.6 & 71.7 & 89.0 & 80.4 & \textbf{93.8} & 77.8 & 91.3 & 66.7 & 98.0 \\ 
        & \multicolumn{2}{c}{Kr = 0.6} & \textbf{84.3} & 78.6 & \textbf{90.7} & 80.4 & 87.5 & 75.6 & \textbf{93.5} & 69.8 & 98.0 \\  
        \midrule
        \multirow{10}{*}{\rotatebox[origin=c]{90}{MIL-EN-B3}} 
        & \multirow{3}{*}{\textbf{I-1}}
        & Max & 74.1 & 57.2 & 81.5 & 74.3 & 77.1 & 74.6 & 77.8 & 61.9 & 88.2 \\
        & & Topk & 78.4 & 72.8 & 78.8 & 75.6 & 79.2 & 74.9 & 87.3 & 68.3 & 90.2 \\ 
        & & Avg & 79.9 & 71.7 & 86.4 & 78.1 & 83.3 & 76.8 & 84.1 & 62.7 & 96.1 \\
        \cmidrule{2-12}
        & \multirow{3}{*}{\textbf{I-2}}
        & Max & 76.4 & 72.3 & 82.4 & 75.4 & 72.9 & 66.7 & 80.0 & 65.1 & 96.1 \\
        & & Topk & 76.2 & 75.7 & 77.1 & 65.5 & 79.2 & 71.0 & 82.5 & 66.7 & 92.2 \\
        & & Avg & 77.5 & 68.2 & 86.5 & 74.1 & 77.1 & 73.5 & 81.2 & 69.0 & 90.2 \\
        \cmidrule{2-12}
        & \multirow{3}{*}{\textbf{E}}
        & Max & 72.3 & 55.5 & 77.6 & 73.7 & 68.8 & 66.9 & 79.0 & 70.6 & 86.3 \\
        & & Topk & 78.9 & 66.5 & 86.3 & 79.6 & 81.3 & 74.2 & 84.1 & 66.7 & 92.2 \\
        & & Avg  & 77.6 & 68.8 & 84.2 & 77.3 & 77.1 & 77.4 & 80.7 & 63.5 & 92.2 \\
        \midrule
        \multirow{10}{*}{\rotatebox[origin=c]{90}{MIL-DEiT-S}} 
        & \multirow{3}{*}{\textbf{I-1}}
        & Max & 82.2 & 72.8 & 88.7 & 81.1 & 89.6 & 80.1 & 82.7 & 64.3 & 98.0 \\
        & & Topk & 81.7 & 78.0 & 84.0 & 73.0 & 66.7 & 78.4 & 88.4 & 68.3 & 90.2 \\ 
        & & Avg & 81.6 & 76.3 & 85.7 & 72.0 & 89.6 & 75.4 & 88.6 & 69.0 & 96.1 \\
        \cmidrule{2-12}
        & \multirow{3}{*}{\textbf{I-2}}
        & Max & 75.4 & 70.5 & 80.9 & 75.8 & 72.9 & 66.6 & 83.1 & 61.1 & 92.2 \\
        & & Topk & 79.0 & 74.0 & 84.5 & 75.6 & 72.9 & 76.3 & 87.2 & 65.1 & 96.1 \\
        & & Avg & 82.6 & 79.2 & 87.8 & 76.2 & 87.5 & 77.9 & 91.1 & 66.7 & 94.1 \\
        \cmidrule{2-12}
        & \multirow{3}{*}{\textbf{E}}
        & Max & 82.4 & \textbf{80.9} & 87.8 & 75.1 & 83.3 & 76.4 & 89.0 & \textbf{74.6} & 92.2 \\
        & & Topk & 82.2 & 73.4 & 83.6 & 74.3 & 93.8 & 75.6 & 92.2 & 69.1 & 96.1 \\
        & & Avg  & 82.6 & 70.5 & 88.7 & 79.8 & 89.6 & 75.1 & 91.5 & 65.9 & \textbf{99.9} \\
        \bottomrule
    \end{tabular}
\end{table*}

\subsection{Additional MIL Heatmaps}

\begin{figure*}[t]
    \centering
    \includegraphics[width=\textwidth]{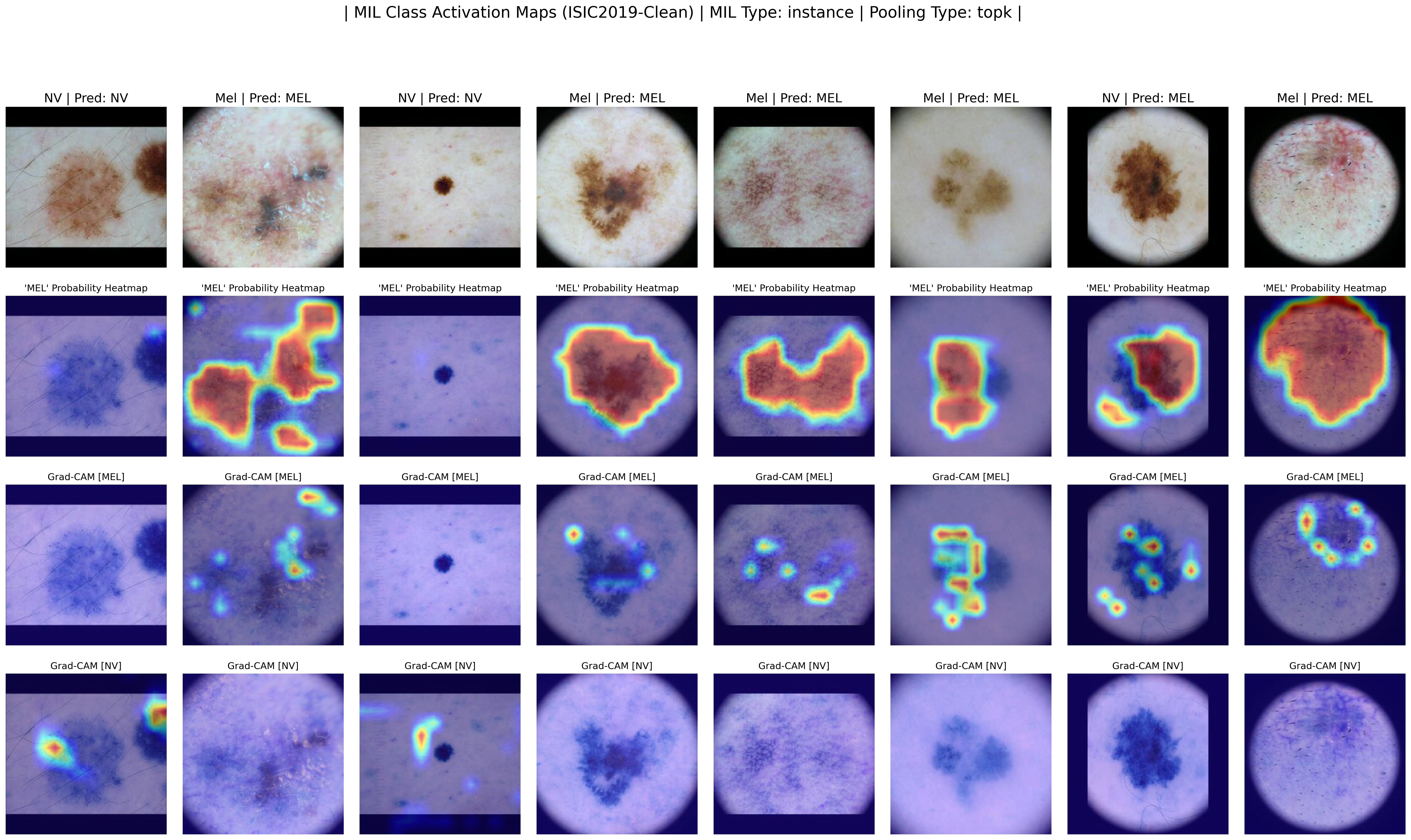}
    \caption{Visualization of the heatmaps generated by the MIL classifier, specifically the \textit{instance-level} MIL model using the \textbf{top-$k$ average pooling} operator. The backbone used for the MIL model is the RN-18. The images are taken from the validation set of the ISIC 2019 dataset, and belong to the binary problem of melanoma vs. nevus. The Figure shows the input images in the first row, followed by the patch probability heatmap for the melanoma class in the second row. The third row shows the gradient heatmap for each patch. In this case, the last row shows the gradients with respect to the patches that the model predicted to be nevus.} 
    \label{fig:appendix_mil_instance_topk_vis}
\end{figure*}

Figure \ref{fig:appendix_mil_instance_topk_vis} provides a visual representation of the different visualizations produced by the \textit{instance-level} MIL model using the \textbf{top-$k$ average pooling} operator. In this case, the last row shows the gradients associated with the patches classified as nevus.

\clearpage